\documentclass[conference]{IEEEtran}
\IEEEoverridecommandlockouts

\usepackage{color}

\usepackage{cite}
\usepackage{url}
\usepackage{amsmath,amssymb,amsfonts}
\usepackage{algorithmic}
\usepackage{graphicx}
\usepackage{textcomp}
\usepackage{xcolor}
\usepackage{multirow}

\usepackage[a4paper, total={184mm,239mm}]{geometry}
\def\BibTeX{{\rm B\kern-.05em{\sc i\kern-.025em b}\kern-.08em
    T\kern-.1667em\lower.7ex\hbox{E}\kern-.125emX}}
    
\usepackage{tikz}
\usepackage{textcomp}
\usepackage[doipre={DOI:~}]{uri}
\usepackage{lipsum}
\newcommand\copyrighttext{%
  \footnotesize \textcopyright 2022 IEEE. Personal use of this material is permitted. Permission from IEEE must be obtained for all other uses, in any current or future media, including reprinting/republishing this material for advertising or promotional purposes, creating new collective works, for resale or redistribution to servers or lists, or reuse of any copyrighted component of this work in other works.
 
  Accepted as a conference paper at the 2022 IEEE International Symposium on Circuits and Systems (ISCAS).}
\newcommand{\copyrightnotice}{%
\begin{tikzpicture}[remember picture,overlay]
\node[anchor=south,yshift=10pt] at (current page.south) {\fbox{\parbox{\dimexpr\textwidth-\fboxsep-\fboxrule\relax}{\copyrighttext}}};
\end{tikzpicture}%
}

\begin{document}

\title{Privacy-preserving Social Distance Monitoring on Microcontrollers with Low-Resolution\\Infrared Sensors and CNNs%
\thanks{This work has received funding from the ECSEL Joint Undertaking (JU) under grant agreement No 101007321. The JU receives support from the European Union’s Horizon 2020 research and innovation programme and France, Belgium, Czech Republic, Germany, Italy, Sweden, Switzerland, Turkey.}
}

\author{\IEEEauthorblockN{Chen Xie\IEEEauthorrefmark{1}, Francesco Daghero\IEEEauthorrefmark{1}, Yukai Chen\IEEEauthorrefmark{1}, Marco Castellano\IEEEauthorrefmark{2}, Luca Gandolfi\IEEEauthorrefmark{2},\\Andrea Calimera\IEEEauthorrefmark{1}, Enrico Macii\IEEEauthorrefmark{1}, Massimo Poncino\IEEEauthorrefmark{1},  Daniele Jahier Pagliari\IEEEauthorrefmark{1}}%
\IEEEauthorblockA{
\IEEEauthorrefmark{1}Politecnico di Torino, Turin, Italy, E-mail: name.surname@polito.it \\
\IEEEauthorrefmark{2}STMicroelectronics, Cornaredo, Italy, E-mail: name.surname@st.com}%
}

\maketitle
\copyrightnotice

\begin{abstract}
Low-resolution infrared (IR) array sensors offer a low-cost, low-power, and privacy-preserving alternative to optical cameras and smartphones/wearables for social distance monitoring in indoor spaces, permitting the recognition of basic shapes, without revealing the personal details of individuals.
In this work, we demonstrate that an accurate detection of social distance violations can be achieved processing the raw output of a 8x8 IR array sensor with a small-sized Convolutional Neural Network (CNN). Furthermore, the CNN can be executed directly on a Microcontroller (MCU)-based sensor node. 

With results on a newly collected open dataset, we show that our best CNN achieves 86.3\% balanced accuracy, significantly outperforming the 61\% achieved by a state-of-the-art deterministic algorithm. Changing the architectural parameters of the CNN, we obtain a rich Pareto set of models, spanning 70.5-86.3\% accuracy and 0.18-75k parameters. Deployed on a STM32L476RG MCU, these models have a latency of 0.73-5.33ms, with an energy consumption per inference of 9.38-68.57$\mu$J.
\end{abstract}

\begin{IEEEkeywords}
Social Distance, COVID-19, Edge Computing, Infrared Sensors, Convolutional Neural Networks
\end{IEEEkeywords}

\section{Introduction and Related Works}\label{sec:introduction}

In the current COVID-19 pandemic, social distancing~\cite{Chanjuan2020,Voko2020,greenstone2020does}, together with extensive testing~\cite{world2020laboratory,Berchialla2021} has been proven the most effective way to prevent the spread of infections, especially when effective treatments and vaccines were not yet available.
This virus spread prevention measure, particularly critical for indoor environments such as offices, shops, factories etc~\cite{Chanjuan2020,Voko2020}, will therefore continue to have an important role in this, and in possible future epidemics. 

Several technical solutions have been proposed to monitor compliance
with social distancing rules. One category is based on computing the distance among people using the Bluetooth or Wi-Fi transcievers available in smartphones~\cite{Rusli2020, Li2021} or wearables~\cite{Cunha2020,Kobayashi2020}. 
While effective, this approach requires the voluntary participation of users,
hence not providing 100\% safety guarantees.
Another set of solutions monitors indoor spaces with cameras~\cite{Ahmed2021,Yang2021,Alhmiedat2021,Jethani2021,Niu2021}. The videos are then processed with Machine Learning (ML) algorithms to locate people and compute their relative distance. While not requiring actions from users, this approach poses critical privacy issues. In fact, workers may have concerns with the installation of a system that can not only monitor the distance among them, but also identify, record and track individuals, often in violation with privacy protection laws~\cite{Barsocchi2021}.

In this scenario, low-resolution infrared (IR) array sensors constitute an interesting alternative. Being capable of acquiring small-sized thermal images (typically 8x8 or 16x16 pixels) at a frame rate of $\approx$10 Frames Per Second (FPS)~\cite{panasonic}, these sensors can recognize basic shapes, without revealing privacy-sensitive details of an individual (facial features, clothes, hair style, etc).
Furthermore, their limited power consumption, low cost, and low-resolution outputs has another positive implication for privacy. That is, 
it enables the implementation of social distance monitoring directly \textit{on the sensing devices},
which are typically battery operated and equipped with resource-constrained Microcontrollers (MCUs)~\cite{Daghero2021,Risso2021,Risso2021a,Gomez2018,Metwaly2019}.
Executing the monitoring on end-nodes, in turn, enables privacy-preserving solutions in which social distance violations are signaled in real time, warning the responsible staff without ever transmitting and/or storing the collected data in the cloud.

The key question is then whether the data collected by low-resolution IR sensors can be effectively used for this task. In fact, while other authors have employed these sensors for various applications~\cite{grideyeapi,mashiyama2015activity,Hayashida2017,tao2018home,Liu2020,Kawashima2017,Gochoo2019,Tao2019,Shih2020,Morawski2020,hanosh2020convulsive,Trofimova2017,Herrmann2018,Gomez2018,Metwaly2019}, to the best of our knowledge no previous work has considered them for social distancing.

Most works combining low-resolution IR sensors with ML focus on human activity recognition~\cite{mashiyama2015activity,Kawashima2017,tao2018home,Gochoo2019,Tao2019,Shih2020,Morawski2020,hanosh2020convulsive,Hayashida2017,Liu2020}. Some of them apply classical algorithms~\cite{mashiyama2015activity,Hayashida2017,tao2018home,Liu2020} while others propose Deep Learning (DL) approaches, based either on Convolutional Neural Networks (CNNs), Long-Short Term Memory (LSTMs), Gated Recurrent Units (GRUs), or a combination of the above (e.g., CNN-LSTM)~\cite{Kawashima2017,Gochoo2019,Tao2019,Shih2020,Morawski2020,hanosh2020convulsive}. Besides the selected algorithm, activity recognition solutions also differ in terms of: i) the number and position of the employed IR sensors, ii) the pre-processing applied to thermal images before feeding them to ML models, and iii) the type of recognized activities, which range from daily tasks such as walking, sitting, standing, etc~\cite{mashiyama2015activity,Kawashima2017,tao2018home,Tao2019,Shih2020,Morawski2020}, to elderly people falls~\cite{Hayashida2017,Liu2020}, epilepsy-induced convulsions~\cite{hanosh2020convulsive}, and even yoga postures~\cite{Gochoo2019}.

Another set of works based on IR thermal arrays focuses on human presence detection~\cite{Trofimova2017,Herrmann2018} and people counting~\cite{grideyeapi,Gomez2018,Metwaly2019}. Also in this case, both classical~\cite{Trofimova2017,grideyeapi} and DL solutions exist~\cite{Herrmann2018,Gomez2018,Metwaly2019}, and the latter are again based on CNNs, LSTMs and GRUs.
In proper conditions, people counting can be transformed into social distance monitoring: if the IR sensor is positioned so that having more than a given number of people in the field of view corresponds to a violation of social distancing rules, then the monitoring algorithm can simply compute the people count and compare it with a threshold to trigger an alarm. However, existing IR array-based people counting implementations use relatively high resolution sensors (e.g., 24x32~\cite{Metwaly2019} and 80x60~\cite{Gomez2018}), which do not guarantee the same level of privacy, and also induce more power consumption than low-resolution ones, both in the sensors themselves and in the computation part of the system. To our knowledge, the only people counting solution for low-resolution IR sensors (8x8 pixels) is the one proposed in~\cite{grideyeapi}, which is not based on ML. However, as shown in Section~\ref{sec:results}, this approach obtains poor results when adapted to social distance monitoring.

In this work, we introduce the \textit{first dedicated implementation} of a privacy-preserving social distance monitoring system on extremely low-resolution IR arrays. We use the same 8x8 sensor of~\cite{grideyeapi}, but we follow an end-to-end deep learning approach based on a CNN classifier.
We explore different CNN architectures, which are then quantized and deployed on a commercial MCU for inference (the STM32L476RG by STMicroelectronics).
With experiments on a new dataset, we show that we can obtain a rich set of Pareto-optimal solutions in the accuracy versus computation cost space. Specifically, all our CNNs meet real-time performance constraints, while reaching 70.5-86.3\% balanced accuracy on the test set, occupying 64-139 kB of Flash memory (including code size) and consuming just 9.38-68.57$\mu$J per inference, depending on the architecture. The balanced accuracy of our best CNN is 25\% higher than the one obtained by the baseline method of~\cite{grideyeapi}.

\section{Proposed Method}
\subsection{Dataset Collection}

While there exist a few public datasets containing IR array sensor images, none of them fits the needs of our target application. In fact, most of them are relative to high resolution sensors (from 160x120 pixels to 640x480), for applications in robotics, autonomous driving, etc~\cite{olmeda2013pedestrian,flir,hwang2015multispectral,Rivadeneira_2020_CVPR_Workshops}. To the best of our knowledge, the only public dataset of low-resolution IR images
is~\cite{baja-1j59-20}. However, the latter is built for more complex human activity recognition tasks. Accordingly, samples are obtained combining the data from 3 different IR sensors, wall-mounted in different positions. In contrast, our goal is to build a social distance monitoring system based on a \textit{single} IR array.

Therefore, we collected a brand new labelled dataset, specifically tailored for person counting and social distance monitoring, which is now available open source\footnote{\url{https://www.kaggle.com/francescodaghero/linaige}}. We decided to use ceiling-mounted sensors, since in indoor environments, the height of the ceiling is less variable than the size of the room. Consequently, the shape of a person seen from above in the thermal image does not change significantly for different environments. As in~\cite{baja-1j59-20}, IR frames are acquired using a Panasonic Grid-EYE (AMG8833), which outputs a 8x8 array, and has a view angle of 60°~\cite{panasonic}.

\begin{figure}[t]
\vspace{-0.3cm}
\centering
\includegraphics[width=.7\columnwidth]{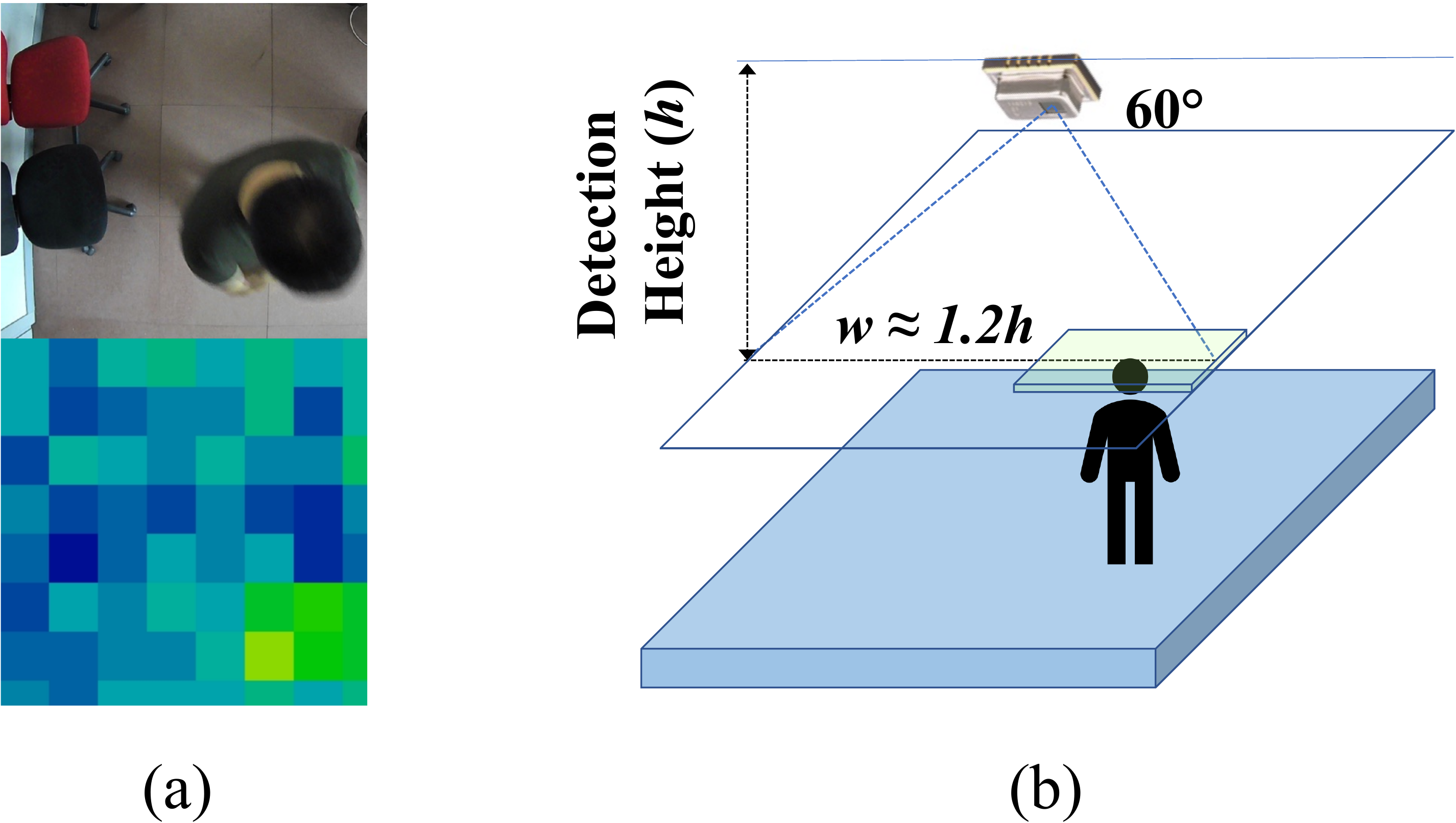}%
\vspace{-0.4cm}
\caption{Example of optical and thermal frames (a) and Panasonic Grid-EYE detecton area calculation (b).}
\vspace{-0.5cm}
\label{fig:detection}
\end{figure}

To gather the data, we setup a system based on a Raspberry Pi 3 equipped with 
the Grid-Eye and a standard USB camera, pointing in the same direction.  We collected synchronized frames from the camera and sensor at 10 FPS, in different indoor environments including offices, research laboratories and corridors, in 6 different sessions, asking up to 5 volunteers to stand, walk, etc, in the area under the sensor.
Table~\ref{tab:dataset} reports the detailed information of each session, whereas Figure~\ref{fig:detection}a shows an example of optical and IR frames.

\begin{table}[t]
    \centering
     \caption{Dataset Information. Abbreviations: Viol = violations.}\label{tab:dataset}
    \scriptsize
    \begin{tabular}{|c|c|c|c|c|c|}
    \hline
    Session & N. & Viol. & N. & Sensor& Room \\
    & Frames & [\%] & People & Height [m] & \\ \hline
    1 & 18124 & 34.3 & 0-5 & 2.4 & A \\ \hline
    2 & 1581 & 54.5 & 0-2 & 2.7 & B \\ \hline
    3 & 1519 & 6.4 & 0-3 & 2.4 & A \\ \hline
    4 & 196 & 5.6  & 0-2 & 2.4 & A\\ \hline
    5 & 2202 & 22.7 & 0-3 & 2.4 & A \\ \hline
    6 & 1850 & 32.3 & 0-3 & 2.7 & C \\ \hline
    \textbf{Tot.} & 25472 & 32.5 & - & - & -\\ \hline
     \end{tabular}
\end{table}

We used a semi-automatic approach to speed-up the labeling of IR data. Namely, we applied a pre-trained Mask R-CNN~\cite{he2018mask} to the optical images, detecting and counting the number of people in each frame, and associating the same people count to the corresponding IR frame. Human labellers were then provided with the two images (optical and thermal) and asked to confirm or correct the people count suggested by Mask R-CNN. Besides coping with Mask R-CNN misclassifications, human checking was also needed because of the difficulty of exactly matching the alignment and viewing angles of the camera and IR array. Due to such imperfect matching, some IR frames showed the heat profile of person (close to a corner) which did not appear on the corresponding optical frame. Human labellers were given the possibility of annotating such ``hard-to-label'' frames, which were then excluded from the training and testing of the proposed model and of the state-of-the-art comparison.

\subsection{Social Distance Monitoring Solution}

Given the view angle of the IR sensor, the width $w$ of the (square) field of view can be calculated with simple trigonometry. As explained in~\cite{panasonic} and shown in Figure~\ref{fig:detection}b, $w \approx 1.2h$ where $h$ is the distance between the sensor and the detected object. For the range of sensor heights present in our dataset (see Table~\ref{tab:dataset}), and assuming that people heads (the main sources of heat detected by the sensor) are at least 1.5m above the floor level, $w$ spans the range
$[1.08:1.44]$ m. Consequently, the field of view diagonal, i.e., the maximum distance between two in-frame objects is in the range $[1.53:2.04]$ m.

\begin{figure}[ht]
\vspace{-0.5cm}
\centering
\includegraphics[width=.7\columnwidth]{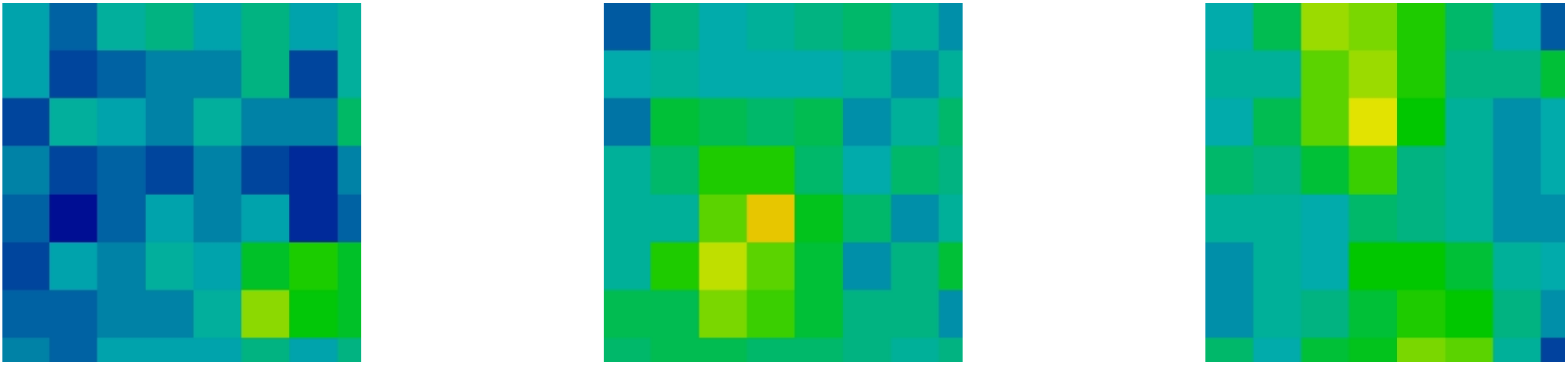} 
\caption{Examples of IR array frames corresponding to 1 person (left), 2 people, close to each other (center), two people, far from each other (right).} \label{fig:example_2}
\vspace{-0.2cm}
\end{figure}

Given that the typical recommendation for social distancing is to maintain at least 2m from the closest person~\cite{Voko2020}, we can conclude that \textit{any frame containing more than 1 person} corresponds to a violation. Accordingly, we model our task as a \textit{binary classification}, where the goal is to predict whether a frame contains 2 or more people. This approach can be extended to 
larger spaces by combining multiple IR arrays placed in different positions on the roof~\cite{panasonic}.
Note that, while this formulation simplifies the problem with respect to predicting the exact people count, it is still significantly more complex than simple presence detection~\cite{Trofimova2017,Herrmann2018}, i.e. distinguishing between no person and 1 or more people. For instance, Figure~\ref{fig:example_2} shows that two nearby people can produce a single hot area in the frame, which can be easily misclassified as a single person.

\begin{figure*}[t]
\centering
\includegraphics[width=.75\textwidth]{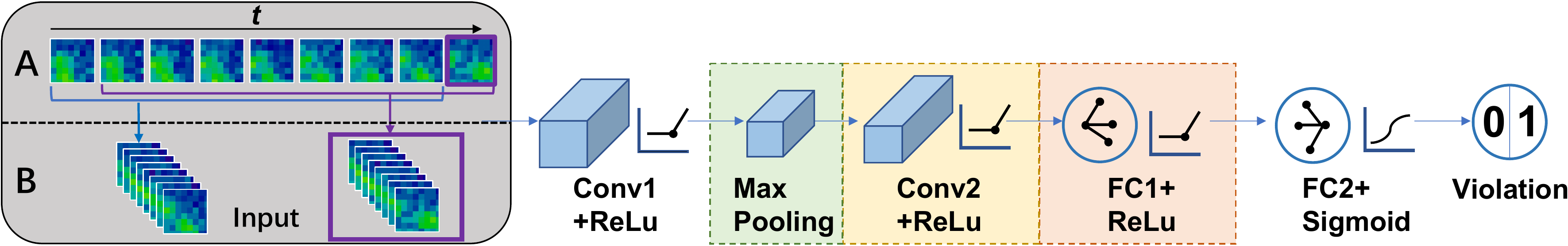} 
\caption{CNN Architecture and input generation.} \label{fig:cnn}
\vspace{-0.3cm}
\end{figure*}

Convolutional Neural Networks (CNNs)~\cite{LeCun1998} are known to be amongst the most effective DL models to analyze visual imagery.
Therefore, we design a set of simple CNN classifiers to implement the proposed IR array-based social distance monitoring. The common template of the models is inspired by classic CNNs such as LeNet-5~\cite{LeCun1998}, and is shown in Figure~\ref{fig:cnn}. It includes two Convolutional (Conv) layers with Rectified Linear Unit (ReLU) activation, one Max Pooling layer, and two Fully Connected (FC) layers. The first FC layer uses a ReLU activation and has a hidden size of 64, while the second one has a sigmoid activation and a single output neuron, producing the probability of a distancing rule violation.

Starting from this template, we perform an extensive architectural exploration. Different networks are obtained eliminating some of the layers enclosed in dashed boxes in Figure~\ref{fig:cnn}. Specifically, we consider architectures with : i) one or two Conv layers, ii) one or two FC layers, iii) with or without Max Pooling. We also vary the number of \textit{channels} in each of the two convolutional layers considering values $\{8, 16, 32, 64\}$. 

We further build two variants of each CNN architecture, differing in the processed input. The first network is fed with a a \textit{single} thermal frame $X_{t}$, and learns to predict whether the corresponding people count $y_{t}$ is greater than 1 (social distance violation). The second variant is fed with a \textit{sliding window} of $W$ consecutive frames ($X_{t},...,X_{t+W-1}$), and is trained to detect social distance violations in the \textit{last} frame of the window (i.e., $y_{t+W-1} > 1$). The corresponding input tensor shapes are (8,8,1) and (8,8,W) respectively, i.e., the $W$ frames are passed to the CNN as different \textit{input channels}. The two types of input are depicted in the left of Figure~\ref{fig:cnn} as A and B respectively. For clarity, a single input sample of each type is enclosed by a purple box.
The second CNN variant has the advantage of having access to \textit{past} frames, which can be used, for instance, to detect that a single heat source corresponds to two nearby people (see Figure~\ref{fig:example_2}), based on their movement. On the other hand, this variant has a higher computational complexity, which is particularly critical for in-sensor inference. In our experiments, we use $W=8$ for a fair comparison with the algorithm of~\cite{panasonic}, which uses a window of 8 frames for background subtraction.

As a further optimization step, to reduce the memory occupation and inference latency/energy of our CNNs, all parameters and intermediate activations are \textit{quantized} to 8-bit integers~\cite{Jacob2018}.

\subsection{Training Protocol}

We trained our CNNs on Session 1 data (see Table~\ref{tab:dataset}) and tested them on all other sessions. This \textit{per-session} split ensures that test samples are taken either in a different room with respect to training data, or in the same room but at a different day/time. In contrast, a purely random split in which consecutive frames from the same session can end up in different data subsets would lead to an overly simplified and unrealistic version of the problem.
Considering all combinations of CNN architectures and input shapes, we trained a total of 96 networks. Each model has been trained 5 times with different random seeds, for a maximum of 500 epochs, with early stopping after 10 non-improving epochs. We used a binary cross-entropy loss function and the Adam optimizer, with an initial Learning Rate (LR) of $1\cdot 10^{-3}$. The LR is reduced by a factor 0.3 on plateau, with a patience of 5 epochs. Once the floating point model reaches convergence, we quantize it and then apply quantization-aware training~\cite{Jacob2018} with the same protocol, except for the initial LR, which is set to $5e^{-4}$.
To deal with the class imbalance of the target dataset (see Table~\ref{tab:dataset}), we apply sample weights to the loss function, equal to the inverse of the class probabilities.
After training all CNNs, we select those that are in the Pareto front as detailed in Section~\ref{sec:results}. %

\section{Results}\label{sec:results}

\begin{table*}[ht]
    \centering
     \caption{Detailed evaluation and deployment Results of three selected architectures on STM32L4 @ 80MHz (12.8 mW average power). Abbreviations: 1/8F = 1-frame/8-frame input, Cn = Convolution with N channels, P = Max Pooling, FC = Fully-Connected.\label{tab:arch}}
    \resizebox{.95\textwidth}{!}{
    \begin{tabular}{|c|c|c|c|c|c|c|c|c|c|c|}
    \hline
      \textbf{Model} & \textbf{Bal. Acc. [\%]} &  \textbf{ROC-AUC} & \textbf{Acc. [\%]}
      & \textbf{F1} & \textbf{Size[B]} & \textbf{Mem.[B]} &\textbf{MACs} &  \textbf{Energy[µJ]} & \textbf{Latency[ms]} & \textbf{Architecture}  \\  \hline
     MinSize & 70.5±3.6 & 0.83±0.04 & 75.6±0.9 & 0.56±0.05 & 0.18k & 63.2k & 2.6k  & 9.38 & 0.728 & 1F-C8-P-FC\\ \cline{1-10}
     
    MaxAcc-1\% & 85.3±2.0 & 0.93±0.04 & 86.4±1.9 & 0.77±0.03 & 4.9k & 68.2k & 7.3k & 12.93 & 1.005 & 1F-C8-P-FC-FC\\ \cline{1-10}
    
    MaxAcc & 86.3±2.2 & 0.99±0.01 & 87.6±1.8 & 0.78±0.03 & 74.7k & 138.8k & 157k & 68.57 & 5.328 & 8F-C32-FC-FC\\ \hline
    
    \cite{grideyeapi} & 61 &-& 73 &0.41&-& 33.5k&- & 53.09 &4.125 &- \\
    
    \hline
     \end{tabular}
     }
     \vspace{-0.4cm}
\end{table*}

We trained and quantized our CNNs with Keras/TensorFlow 2.0~\cite{tensorflow}, and we used the X-CUBE-AI v6.0 toolchain~\cite{cubeai} to convert trained models into optimized C code for inference. Given the class imbalance of the target dataset, we mainly evaluate the performance of our models in terms of \textit{balanced accuracy} (Bal. Acc.), i.e. the arithmetic mean of sensitivity and specificity.
However, we also report the Area under the ROC (ROC-AUC), the F1-Score (F1), and the standard micro-average accuracy (Acc.)~\cite{statistics_book}.
For all metrics, we report mean $\pm$ standard deviation on the 5 training runs.

The hardware-independent computational cost of the CNNs, used to extract Pareto curves, is measured in terms of model size (number of parameters) and number of Multiply-and-Accumulate (MAC) operations. Models deployed on the MCU are then further evaluated in terms of inference latency, energy consumption, and memory occupation.

We mainly compare against the algorithm of~\cite{grideyeapi}, which is, to our knowledge, the only person counting solution on low-resolution IR sensors. We execute the algorithm of~\cite{grideyeapi} as is, and simply convert the person count into a social distance violation whenever 2 or more people are detected in a frame.

\subsection{Architectural Exploration}

\begin{figure}[ht]
\centering
\includegraphics[width=0.8\columnwidth]{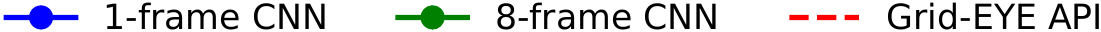}
\includegraphics[width=.8\columnwidth]{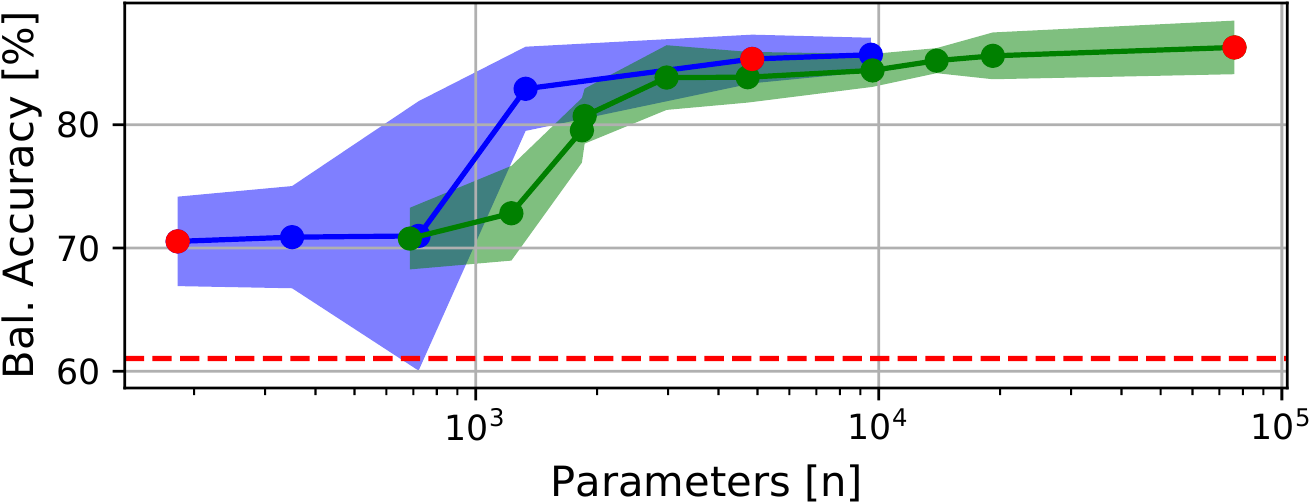}
\includegraphics[width=.8\columnwidth]{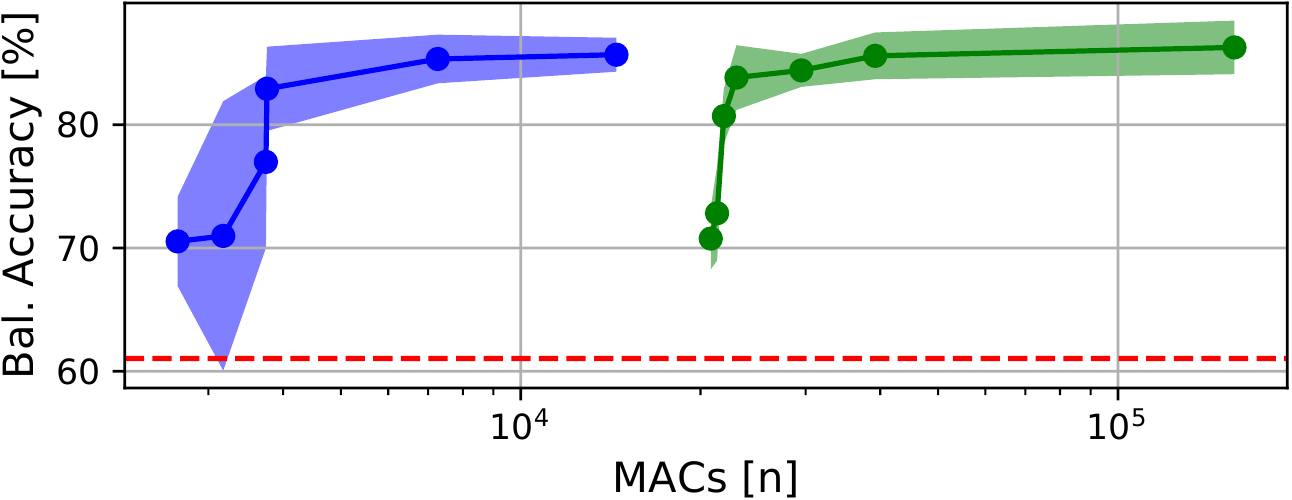}
\vspace{-0.4cm}
\caption{Pareto Curves of balanced accuracy vs. parameters and n. of operations.}\label{fig:curves}
\vspace{-0.4cm}
\end{figure}

Fig. \ref{fig:curves} shows the Pareto curves generated by different CNN architectures in terms of balanced accuracy versus number of parameters and MAC operations respectively. The two curves refer to the single-frame and windowed input variants respectively (A and B in Figure~\ref{fig:cnn}), and each point represents a different architectural configuration (type and number of layers, number of channels, etc). Dots correspond to the mean accuracy over 5 trainings, while colored bands identify the $\pm$ standard deviation range. The horizontal dashed line shows the accuracy obtained by the baseline method of~\cite{grideyeapi}. Note that the model size and MAC curves contain in general \textit{different} CNNs.

The results clearly show that all proposed CNNs significantly outperform the baseline comparison in terms of accuracy. In particular, the single-frame networks achieve very high accuracy ($>80$\%) even with just around 1k parameters and less than 10k MACs. Windowed networks are significantly more costly, especially in terms of number of operations, yet they are able to obtain the overall best accuracy (86.3\%), in a configuration with 75k parameters, and requiring 157k MACs.

In general, we found that both types of models achieve better performance using a small number of channel in convolutional layers (architectures with 64 channels are not in the Pareto front). This is probably due to the relatively small and simple dataset, for which large number of features lead to over-fitting.

\subsection{Deployment}

We selected 3 architectures from the parameters Pareto curve (indicated by red dots in Figure~\ref{fig:curves}) and deployed them on the target MCU. The detailed results are shown in Table \ref{tab:arch}. The selected CNNs are: i) the most accurate one, ii) the smallest model that achieves a $<1\%$ balanced accuracy drop with respect to the best, and iii) the smallest overall. We also compiled and profiled the code of~\cite{grideyeapi} on the same MCU for comparison. For our CNNs, the table reports both the quantized model size in Bytes (Size column), as well as the total occupied Flash memory, also including code size (Mem). As shown, our CNNs outperform~\cite{grideyeapi} in all considered accuracy metrics. Note that the ROC-AUC is not reported for~\cite{grideyeapi} since this method outputs a deterministic 0/1 value, and not a probability score. In terms of bal. acc., we outperform~\cite{grideyeapi} by 9.5-25.2\%, depending on the selected CNN. Moreover, MinSize and MaxAcc-1\% are respectively 5.6x and 4.1x faster and more energy efficient than~\cite{grideyeapi}. Although the total memory occupation of our models is larger, this is mostly due to the large code size of X-CUBE-AI ($\approx$60kB). Moreover, all three CNNs easily fit in the 1MB Flash of the target MCU.
Finally, note that, mainly because we rely on a very low-resolution 8x8 sensor, our MinSize/MaxAcc CNNs are 60.6x/8.3x faster than the networks proposed in~\cite{Metwaly2019}, and strikingly 235000x/32000x more energy efficient than the method of~\cite{Gomez2018}, i.e., the only other two works proposing MCU deployments of IR array-based ML tasks. However, it must be underlined that, besides using different inputs, \cite{Metwaly2019,Gomez2018} also solve a slightly different task (person counting), making a fair comparison with our work impossible.

\section{Conclusions}
We have proposed a new method to implement privacy-preserving social distance monitoring in indoor spaces, using a low-resolution IR array sensor and an end-to-end deep learning approach. Our results show that, with compact and energy-efficient CNNs, an accurate social distance monitoring can be implemented directly on MCU-based sensor nodes. Future works will include the application of sub-byte quantization to further reduce models sizes and complexity.

\bibliographystyle{IEEEtran}

\end{document}